\documentclass[runningheads]{llncs}

\usepackage[T1]{fontenc}
\usepackage{graphicx}
\usepackage{amsmath}
\usepackage{amssymb}
\usepackage{algorithm}
\usepackage{algorithmic}
\usepackage{booktabs}
\usepackage{multirow}
\usepackage{xcolor}
\usepackage{subcaption}
\usepackage{threeparttable}
\usepackage{tikz}
\usepackage{marvosym}
\usepackage{hyperref}
\usepackage{orcidlink} 
\usepackage{etoolbox}

\begin{document}

\title{Tell Me Why: Designing an Explainable LLM-based Dialogue System\\
for Student Problem Behavior Diagnosis}
\titlerunning{Explainable LLM-based Dialogue System for Student Problem Behavior}

\author{
Zhilin Fan\inst{1}\orcidlink{0009-0007-2832-9595} \and
Deliang Wang\inst{3}\orcidlink{0009-0008-6488-0234} \and Penghe Chen\inst{1,2} \and Yu~Lu\textsuperscript{1,2(\Letter)}\orcidlink{0000-0003-2378-4971}
}

\authorrunning{Z. Fan et al.}

\institute{School of Educational Technology, Beijing Normal University, Beijing, China \and Advanced Innovation Center for Future Education, \\ Beijing Normal University, Beijing, China \\ \email{luyu@bnu.edu.cn} \and Faculty of Education, The University of Hong Kong, Hong Kong SAR, China }

\maketitle
\setcounter{footnote}{0}
        \begin{abstract}
            Diagnosing student problem behaviors requires teachers to synthesize multifaceted information, identify behavioral categories, and plan intervention strategies. Although fine-tuned large language models (LLMs) can support this process through multi-turn dialogue, they rarely explain why a strategy is recommended, limiting transparency and teachers' trust. To address this issue, we present an explainable dialogue system built on a fine-tuned LLM. 
            The system uses a hierarchical attribution method based on explainable AI (xAI) to identify dialogue evidence for each recommendation and generate a natural-language explanation based on that evidence.
            In technical evaluation, the method outperformed baseline approaches in identifying supporting evidence. In a preliminary user study with 22 pre-service teachers, participants who received explanations reported higher trust in the system. These findings suggest a promising direction for improving LLM explainability in educational dialogue systems.
            \end{abstract}

\section{Introduction}
Student problem behaviors, such as aggression and rule breaking, can negatively affect students' psychological well-being and academic development~\cite{sutherland2020preliminary}. Diagnosing these behaviors requires teachers to synthesize information about the student, family, and school context, identify behavioral categories, and plan interventions~\cite{chen2025knowstu}. This process is challenging because it draws on knowledge from multiple domains, including psychology and pedagogy~\cite{chen2024pbchat}.

Recent studies suggest that fine-tuned large language models (LLMs) can support student problem behavior diagnosis through multi-turn dialogue and generate intervention strategies more consistent with expert practice~\cite{chen2024pbchat,chen2025knowstu}. However, these systems typically recommend strategies without explaining why they are appropriate. Because teachers are professionally accountable for the strategies they choose~\cite{fan2025explainable}, this lack of explanation may limit their trust in and use of the recommendations.

To address this gap, we present an explainable dialogue system built on a fine-tuned LLM. For each recommended strategy, the system identifies teacher-provided dialogue evidence and generates a natural-language explanation grounded in that evidence. Building on recent explainable AI (xAI) attribution methods for LLMs~\cite{qian2026actionunveilinginternaldrivers,chuang2025selfcite}, we adopt a hierarchical attribution method that first locates the most influential dialogue turn and then identifies the teacher sentence within that turn. We address two research questions (RQs):
\begin{quote}
    \textbf{RQ1.} To what extent can the hierarchical attribution method identify dialogue evidence that supports recommended strategies?

    \textbf{RQ2.} How do explanations influence teachers' trust in the explainable diagnostic dialogue system?
\end{quote}

\section{System Design and Implementation}

\subsection{System Design}
\begin{figure}[!htpb]
    \centering
    \includegraphics[width=1\linewidth]{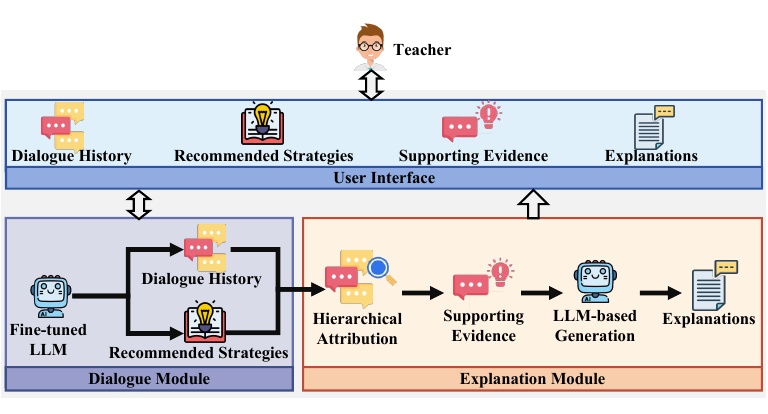}
    \caption{Overview of the explainable diagnostic dialogue system.}
    \label{fig:overview}
\end{figure}

Figure~\ref{fig:overview} summarizes the explainable diagnostic dialogue system, which consists of a \textbf{dialogue module} for multi-turn diagnosis and intervention strategy recommendation, and an \textbf{explanation module} for identifying dialogue evidence and generating natural-language explanations.
\vspace{-10pt}

The system is shaped by two design decisions. First, because general-purpose LLMs lack the proactive diagnostic inquiry capabilities required for student problem behavior diagnosis~\cite{chen2025knowstu}, the dialogue module is built on a fine-tuned LLM that can proactively collect and synthesize multifaceted student information from teachers. Second, because self-generated LLM explanations may sound plausible without faithfully reflecting the basis of a recommendation~\cite{turpin2023language}, the explanation module first identifies dialogue evidence via xAI attribution and then prompts the fine-tuned LLM to generate a natural-language explanation.

\subsection{Dialogue Module Implementation}
We fine-tuned Qwen2.5-3B-Instruct on the training set of the expert-annotated diagnostic dialogue corpus (3,636 examples) from Chen et al.~\cite{chen2025knowstu} using LoRA\footnote{\url{https://huggingface.co/docs/peft}}, with hyperparameters selected via five-fold cross-validation. On the corresponding test set (409 examples)~\cite{chen2025knowstu}, the fine-tuned LLM achieved a macro-F1 of 0.71 for problem behavior type identification and a BERTScore of 0.79 for generated intervention strategies compared with authentic practices, providing a reasonable basis for the explanation module. Additional implementation details are available on GitHub\footnote{\url{https://github.com/zhilinfan/AIED2026-Explainable-Dialogue-System}}.

\subsection{Explanation Module Implementation}

\subsubsection{Turn-level attribution.}
Drawing on temporal likelihood dynamics from prior work~\cite{qian2026actionunveilinginternaldrivers},
we first identify the most relevant dialogue turn to efficiently narrow down the evidence search in long multi-turn dialogues.
Let $r_l$ denote the recommended intervention strategy and $C_i$ the dialogue prefix up to turn $i$.
We quantify the contribution of each turn by comparing the model's support toward $r_l$ with and without the turn:
\begin{equation}
    g_i =
    \underbrace{\log P_{\theta}(r_l \mid C_i)}_{\text{with turn } i}
    -
    \underbrace{\log P_{\theta}(r_l \mid C_{i-1})}_{\text{without turn } i},
    \quad
    i^* = \arg\max_i g_i .
\end{equation}
A large $g_i$ indicates that adding turn $i$ substantially increases the model's support for the recommended intervention strategy. We therefore select the turn with the largest $g_i$, denoted $i^*$, as the most influential turn for subsequent sentence-level attribution.

\subsubsection{Sentence-level attribution.}\label{sec:sent-attr}
Within this turn, we focus on teacher utterances,
which serve as the core diagnostic context,
denoted $u_{i^*} = \{s_{i^*,j}\}_{j=1}^{n_{i^*}}$. Following recent context attribution work~\cite{cohen2024contextcite,chuang2025selfcite}, we score each sentence using two complementary signals: necessity and sufficiency. 

The first signal, \emph{Drop}, captures \textbf{necessity} via a leave-one-out (LOO) ablation~\cite{cohen2024contextcite}. If removing a sentence from the teacher-provided context $U$ lowers the likelihood of the recommended intervention strategy, the model is likely relying on that sentence as supporting evidence:
\begin{equation}
    \text{Drop}(s_{i^*,j}) =
    \underbrace{\log P_{\theta}(r_l \mid U)}_{\text{full context}}
    -
    \underbrace{\log P_{\theta}(r_l \mid U \setminus s_{i^*,j})}_{\text{without sentence } j},
\end{equation}
where $U$ denotes the teacher-provided context in the selected turn.

The second signal, \emph{Hold}, captures \textbf{sufficiency}. If a sentence on its own still preserves support for the recommended intervention strategy, it contains standalone evidence rather than relying only on surrounding context:
\begin{equation}
    \text{Hold}(s_{i^*,j}) =
    \underbrace{\log P_{\theta}(r_l \mid s_{i^*,j})}_{\text{sentence } j \text{ alone}}
    -
    \underbrace{\log P_{\theta}(r_l \mid U)}_{\text{full context}}.
\end{equation}

Following~\cite{chuang2025selfcite}, we combine the two signals by addition so that sentences are favored when they are both necessary in context and informative on their own:
\begin{equation}
\phi_{i^*,j} = \text{Drop}(s_{i^*,j}) + \text{Hold}(s_{i^*,j})
= \underbrace{\log P_{\theta}(r_l \mid s_{i^*,j})}_{\text{sentence } j \text{ alone}} - \underbrace{\log P_{\theta}(r_l \mid U \setminus s_{i^*,j})}_{\text{without sentence } j}.
\end{equation}
A sentence therefore receives a high score when it supports the recommended intervention strategy better on its own than the remaining context does without it. 
We select the top-ranked sentence as supporting evidence for the following explanation generation for teachers.

\begin{figure}[htpb]
    \centering
    \includegraphics[width=1.0\linewidth]{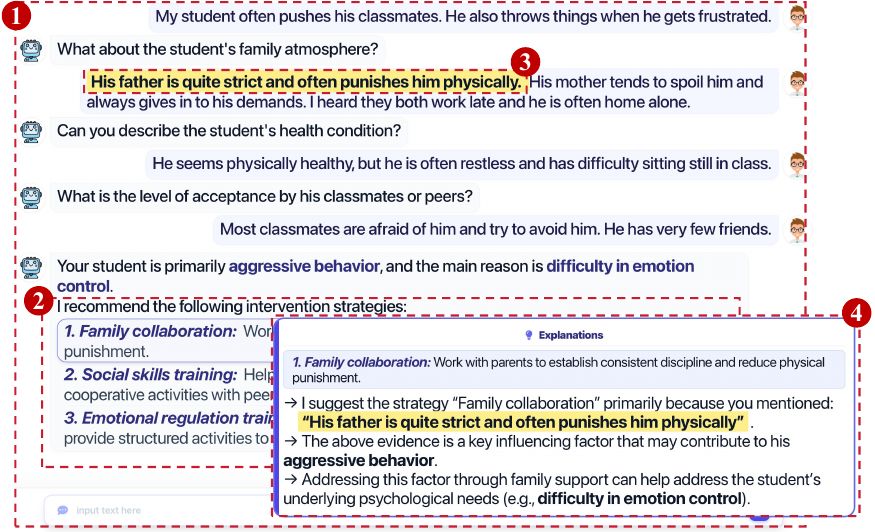}
    \caption{System interface showing (1) the dialogue history, (2) recommended strategies, (3) the identified supporting evidence sentence, and (4) the natural-language explanation.}
    \label{fig:webui}
\end{figure}

Because raw attribution scores are difficult for teachers to interpret, we convert the selected evidence into a natural-language explanation. Specifically, the system prompts the same fine-tuned LLM to generate an explanation conditioned on the recommended intervention strategy and the selected evidence sentence. Fig.~\ref{fig:webui} shows a snapshot of the user interface.


\section{Evaluation and Results}\subsection{RQ1: Identifying Supporting Dialogue Evidence}
We constructed an evaluation benchmark on the test set from Chen et al.~\cite{chen2025knowstu}. Two educational experts annotated teacher-uttered sentences that support each recommended strategy, achieving substantial inter-annotator agreement (Cohen’s $\kappa=0.81$). Disagreements were resolved through discussion, yielding a reliable ground truth.

We compared the hierarchical attribution method against four baselines that directly score all teacher utterances across the dialogue: Drop+Hold, Leave-one-out~\cite{cohen2024contextcite}, GradNorm, and Similarity. Following prior work~\cite{qian2026actionunveilinginternaldrivers}, we report Hit@1, Hit@3, Hit@5, and MRR.
\begin{table}[!htpb]
    \centering
    \caption{Comparison of attribution methods for identifying dialogue evidence.}
    \label{tab:attr}
    \begin{tabular*}{0.8\linewidth}{@{\extracolsep{\fill}}lcccc@{}}
        
    \toprule
    Method & Hit@1 & Hit@3 & Hit@5 & MRR \\
    \midrule
    \textbf{Hierarchical (Ours)} & \textbf{0.778} & \textbf{0.856} & \textbf{0.945} & \textbf{0.803} \\
    Drop+Hold & 0.736 & 0.825 & 0.911 & 0.766 \\
    Leave-one-out & 0.641 & 0.724 & 0.850 & 0.719 \\
    GradNorm & 0.578 & 0.701 & 0.804 & 0.686 \\
    Similarity & 0.511 & 0.670 & 0.738 & 0.639 \\
    \bottomrule
    \end{tabular*}
    \end{table}
As shown in Table~\ref{tab:attr}, the hierarchical attribution method outperforms all baselines across metrics. 
Compared to using Drop+Hold alone without turn-level attribution, it improves Hit@1 from 0.736 to 0.778, suggesting the value of narrowing the search to the most influential turn. Drop+Hold also exceeds Leave-one-out, indicating the benefit of incorporating the sufficiency signal. 
The Hit@1 of 0.778 indicates that the top-ranked sentence matches the ground-truth evidence in over 77\% of cases, 
supporting our design choice of selecting the top-1 sentence for explanation generation (Section~\ref{sec:sent-attr}).

\subsection{RQ2: Preliminary Trust Study}
To examine how explanations relate to teachers' trust, we conducted a preliminary study with 22 pre-service teachers (mean age $= 23.9$, SD $= 1.4$), randomly assigned to a treatment group ($n = 12$) or a control group ($n = 10$). All participants first used the baseline system without explanations to diagnose five student cases and then completed a six-item trust scale in a five-point Likert format~\cite{trustscale}. Next, both groups diagnosed five comparable cases: the treatment group used the explainable system, whereas the control group continued to use the baseline system. The trust questionnaire showed good internal consistency (Cronbach's $\alpha = 0.81$ at pre-test and $0.77$ at post-test). Because the trust scores were ordinal and the sample was small, we used non-parametric tests: the Wilcoxon signed-rank test for within-group pre--post comparisons and the Mann-Whitney $U$ test for between-group comparisons.

    

Descriptively, the treatment group's median trust increased from 17 (IQR 17--18) to 19 (IQR 18--20), whereas the control group remained at 18 (IQR 17--19 to 16--19). Baseline trust did not differ significantly between groups ($U = 50.0$, $p = .514$). In the treatment group, trust increased from pre- to post-test ($W = 3.5$, $p = .012$), whereas the control group showed no significant change ($W = 10.0$, $p > .05$).
A Mann-Whitney test on change scores indicated a between-group difference ($U = 93.5$, $p = .026$, $r = .56$), with larger trust gains in the explanation condition.

\section{Conclusions and Future Directions}

This study explores how to make LLM-based dialogue systems more explainable for teachers in the context of student problem behavior diagnosis. We present an explainable dialogue system, combining a dialogue module built on a fine-tuned LLM with an explanation module based on an xAI attribution method. For each recommended intervention strategy, the explanation module identifies the most influential dialogue turn, ranks teacher-uttered sentences within that turn by necessity and sufficiency, and finally uses the fine-tuned LLM to generate a natural-language explanation grounded in the selected evidence. We evaluate the approach through both a technical evaluation and a preliminary user study with 22 pre-service teachers.

Results suggest that the hierarchical attribution method performs better than baseline approaches in identifying dialogue evidence associated with recommended strategies. Findings from the user study further suggest that providing explanations may improve teachers' trust in the system. This pattern is broadly consistent with prior work on xAI-based explanations for teachers~\cite{wang,Cukurovafeldman2025impact,wang2024making}. Together, these findings suggest that grounding explanations in dialogue evidence using xAI attribution methods may be a promising direction for developing more trustworthy LLM-based educational dialogue systems.

However, several limitations should be noted. The user study involved a small sample of pre-service teachers and measured only self-reported trust, so the findings do not establish whether explanations improve downstream decision making or intervention quality. In addition, the current system explains only recommended intervention strategies, and the attribution method focuses on a single most influential dialogue turn, which may overlook evidence distributed across multiple turns. Future work should therefore evaluate the system with larger and more diverse teacher populations, extend explanation coverage to other diagnostic components, explore multi-turn and richer explanation forms such as counterfactual explanations~\cite{khosravi2022explainable}, and examine generalizability to the full corpus and larger LLMs.

\section*{Acknowledgments}
This research is supported by the National Natural Science Foundation of China (No.62477003, No.62177009).

\bibliographystyle{splncs04}
\bibliography{ref}

\end{document}